\documentclass{article}
\usepackage{ijcai25}

\usepackage{times}
\usepackage{soul}
\usepackage{url}
\usepackage[hidelinks]{hyperref}
\usepackage[utf8]{inputenc}
\usepackage[small]{caption}
\usepackage{graphicx}
\usepackage{amsmath}
\usepackage{amsfonts}
\usepackage{amsthm}
\usepackage{booktabs}
\usepackage{algorithm}
\usepackage{algorithmic}
\usepackage[switch]{lineno}
\usepackage{multirow}
\usepackage{xcolor}
\usepackage{booktabs} 
\usepackage{graphicx}
\usepackage{textcomp}
\usepackage{hyperref}
\usepackage{pifont}
\usepackage{float}
\usepackage{diagbox}
\usepackage{enumitem}
\usepackage{amsmath}
\usepackage{CJKutf8}
\usepackage[T1]{fontenc}

\newcommand{\stitle}[1]{\vspace*{0.4em}\noindent{\bf #1.\/}}
\newcommand{\squishlist}{
	\begin{list}{$\bullet$}
		{ \setlength{\itemsep}{1pt}
			\setlength{\parsep}{1pt}
			\setlength{\topsep}{2.5pt}
			\setlength{\partopsep}{0.5pt}
			\setlength{\leftmargin}{1em}
			\setlength{\labelwidth}{1em}
			\setlength{\labelsep}{0.6em}
		}
	}
	\newcommand{\squishend}{
	\end{list}
}

\title{Guiding LLM-based Smart Contract Generation with Finite State Machine}

\author{
Hao Luo$^{1,\dagger}$\and
Yuhao Lin$^{1,\dagger}$\and
Xiao Yan$^1$\and
Xintong Hu$^2$\and
Yuxiang Wang$^1$\and
Qiming Zeng$^1$\and
Hao Wang$^{1,\ast}$\And
Jiawei Jiang$^{1,\ast}$\\
\affiliations
$^1$School of Computer Science, Wuhan University\\
$^2$School of Cyber Science and Engineering, Wuhan University \\
\emails
\{lohozz, yuhao\_lin\}@whu.edu.cn,
yanxiaosunny@gmail.com,
\{xintong.hu, nai.yxwang, kirin\_z, wanghao.cs, jiawei.jiang\}@whu.edu.cn
}

\begin{document}
\begin{CJK}{UTF8}{gbsn}

\maketitle

\renewcommand{\thefootnote}{\relax}

\footnotetext{$^\dagger$Equal Contribution}
\footnotetext{$^\ast$Corresponding Author}

\begin{abstract}
Smart contract is a kind of self-executing code based on blockchain technology with a wide range of application scenarios, but the traditional generation method relies on manual coding and expert auditing, which has a high threshold and low efficiency. Although Large Language Models (LLMs) show great potential in programming tasks, they still face challenges in smart contract generation w.r.t. effectiveness and security. To solve these problems, we propose FSM-SCG, a smart contract generation framework based on finite state machine (FSM) and LLMs, which significantly improves the quality of the generated code by abstracting user requirements to generate FSM, guiding LLMs to generate smart contracts, and iteratively optimizing the code with the feedback of compilation and security checks.
The experimental results show that FSM-SCG significantly improves the quality of smart contract generation. Compared to the best baseline, FSM-SCG improves the compilation success rate of generated smart contract code by at most 48\%, and reduces the average vulnerability risk score by approximately 68\%.
\end{abstract}
\section{Introduction}
Blockchain, with its characteristics of immutability, transparency, and decentralization, has demonstrated a wide range of applications in various fields such as finance, supply chain, data sharing, etc \cite{wang2018overview}. 
As a key component for realizing blockchain systems, the automatic generation of smart contracts has become one of the focuses in current research \cite{khan2021blockchain}. 

Smart contracts require precise coding and verification to ensure their effectiveness and security~\cite{almakhour2020verification}.
Due to the irreversible nature of smart contract execution, any vulnerabilities or errors in the code could lead to serious security risks, or even financial losses~\cite{luu2016making,sayeed2020smart}.
Therefore, the ability to quickly and reliably generate fully functional and secure smart contracts has become a critical issue.
Previously, the generation of smart contracts primarily relies on manual coding and professional audits of experts, requiring developers to have extensive blockchain programming skills and  experience~\cite{mao2019visual}. 

In recent years, LLMs have demonstrated significant potential in programming tasks, such as code generation, program repair, and code translation~\cite{poesia2022synchromesh,jiang2024survey}. 
Motivated as such, LLMs have also shifted the paradigm of smart contract generation.
Existing works on LLMs for generating smart contracts mainly focus on direct generation from user requirements to code (R2C)
~\cite{napoli2024leveraging,chatterjee2024efficacy}. 
Other works transform requirements into intermediate models to guide LLMs in code generation (requirements-to-model-to-code, R2M2C)~\cite{wohrer2020domain,petrovic2023model,qasse2023chat2code}.

As shown in Table \ref{tab:table-1}, when the LLMs directly generate smart contracts based on user requirements, the code has a low compile pass rate (CPR) and a high vulnerability risk score (VRS). 
When intermediate states, such as the Contract Modeling Language (CML), IContractML, and the FSM in our proposed method, are introduced to guide the LLMs to generate smart contracts, all metrics significantly improve.

\setlength{\tabcolsep}{3pt}
\begin{table}
\centering
\renewcommand{\arraystretch}{1.05}
\caption{
Comparing the existing smart contract generation methods. 
FSM-SCG stands for our proposed approach, while FSM refers to the variant that only uses the proposed SmartFSM as an intermediate representation to aid smart contract generation.
Higher CPR indicates better effectiveness, and lower VRS means better security. We evaluate the methods on the LlaMa3.1-8b model.
}
\vspace{-5pt}
\footnotesize
\label{tab:table-1}
\begin{tabular}{cccc}
\hline
Method & Pattern & CPR(↑) & VRS(↓) \\ \hline
Direct & R2C & 36.9\% & 7.44 \\ 
CML & R2M2C & 47.5\% & 6.36 \\ 
IContractML & R2M2C & 46.3\% & 6.21 \\ 
FSM (Ours) & R2F2C & 49.3\% & 6.05  \\ 
FSM-SCG (Ours) & R2F2C & \textbf{95.1\%} & \textbf{2.36} \\ \hline
\end{tabular}
\vspace{-10pt}
\end{table}

\stitle{Challenges}
Despite pioneering efforts to generate smart contracts using LLMs, there are notable limitations concerning their essential features.
\squishlist
\item Effectiveness: Smart contracts typically use low-resource programming languages, such as Solidity, Vyper, and Move, which lack sufficient documentation, libraries, and community support.
Consequently, in the internal representation space of LLMs, the distribution of low-resource programming languages is often more concentrated compared to high-resource programming languages, such as Python or JavaScript~\cite{li2024quantifying}.
This disparity leads to a higher occurrence of syntax errors and incomplete logic in generated smart contracts, resulting in low compilation success rates.

\item Security: 
Existing research~\cite{chatterjee2024efficacy,jiang2024detecting} has found that LLMs frequently overlook security issues when generating smart contract code.
This oversight stems from the use of general-purpose LLM services (e.g., ChatGPT, OpenAI Codex), which are not specifically designed for the blockchain domain, potentially introducing security vulnerabilities in the generated contracts.

\squishend

To address these limitations, we try to study the following question in this work ---
{\em How to leverage LLMs to generate effective and secure smart contracts?}

\stitle{Our solution FSM-SCG}
Formal methods enable precise modeling of requirements and systems to ensure high reliability and security. 
FSM, as a formal method, can capture the state changes and condition triggers in smart contract to guide secure and reliable generation\cite{suvorov2019smart}.
Therefore, we consider incorporating FSM into the smart contract generation process. 
Additionally, inspired by the Chain-of-Thought (CoT)\cite{wei2022chain} and Structured Chain-of-Thought (SCoT)\cite{li2023structured} techniques, we divide the smart contract generation process into two subtasks: generating FSMs from user requirements and generating smart contracts from FSMs. 
Our approach follows a requirement-to-FSM-to-code (R2F2C) generation pattern, guided by FSM throughout the process, which we call FSM-SCG. 
The workflow begins by fine-tuning LLM to enhance its ability to generate FSMs from user requirements, and then generate smart contracts from FSMs. 
Then, we format user requirements, with which the fine-tuned LLMs generate FSMs.
After this, the generated FSMs undergo format and graph checks. Once verified, FSMs are input into LLMs to generate the corresponding smart contract code. The resulting contracts are evaluated in terms of effectiveness and security.
These validations are utilized as feedback to refine the generation results.

To summarize, we make the following contributions.

\squishlist
    \item 
    We propose a smart contract generation framework called FSM-SCG upon LLMs.
    By abstracting user requirements as FSM, FSM-SCG can significantly improve the generation of smart contracts.

    \item 
    We construct a fine-tuning dataset to improve the ability of LLMs to generate FSMs and smart contracts.
    Additionally, we conduct compilation and security checks, and use them as feedback to optimize the generation of smart contracts.

    \item 
    Experimental results show that FSM-SCG significantly enhances the effectiveness and security of smart contracts. Using LlaMa3.1-8B, the compilation success rate reaches 95.1\%, 48\% higher than the best baseline. Security risks are greatly reduced, with the vulnerability risk score dropping by 68\% on average.
    
\squishend
\section{Preliminaries}

\subsection{Smart Contract Generation Process}

There are two approaches to generating smart contracts using LLMs. The first is the direct generation approach, which inputs user requirements directly for the LLM to generate corresponding smart contract code. The second is the intermediate representation approach, where the LLM first transforms user requirements into an abstract model, then generates smart contract code from this model, effectively decomposing the task into two steps.
The direct generation approach prioritizes rapid implementation and user interaction, while the intermediate representation approach focuses on modeling and abstraction, making it better suited for complex requirements.  
Our work adopts the second approach, utilizing FSM as the intermediate representation.

\subsection{Finite State Machine}
\label{FSM-intro}

FSM is an abstract computational model used to represents the dynamic behavior of a system by recognizing input sequences and transitioning between states according to predefined rules.

A Mealy state machine is a type of FSM. 
Its main characteristic is that the output depends not only on the current state but also on the input signal, allowing for an instantaneous and precise response when the input changes. Typically, a Mealy state machine can be defined as a quintuple $(S, X, Y, \delta, \lambda)$:

\squishlist
    \item $S$ represents the set of states;
    \item $X$ represents the set of inputs;
    \item $Y$ represents the set of outputs;
    \item $\delta: S \times X \rightarrow S$ is the state transition function, which defines the transition rules under different inputs;
    \item $\lambda: S \times X \rightarrow Y$ is the output function that determines the system output based on the current state and input.
\squishend

The Mealy state machine, although it can represent the state transfer of smart contracts well, has limited guidance for the contract generation task. 
\begin{figure*}[t]
    \includegraphics[width=0.95\textwidth]{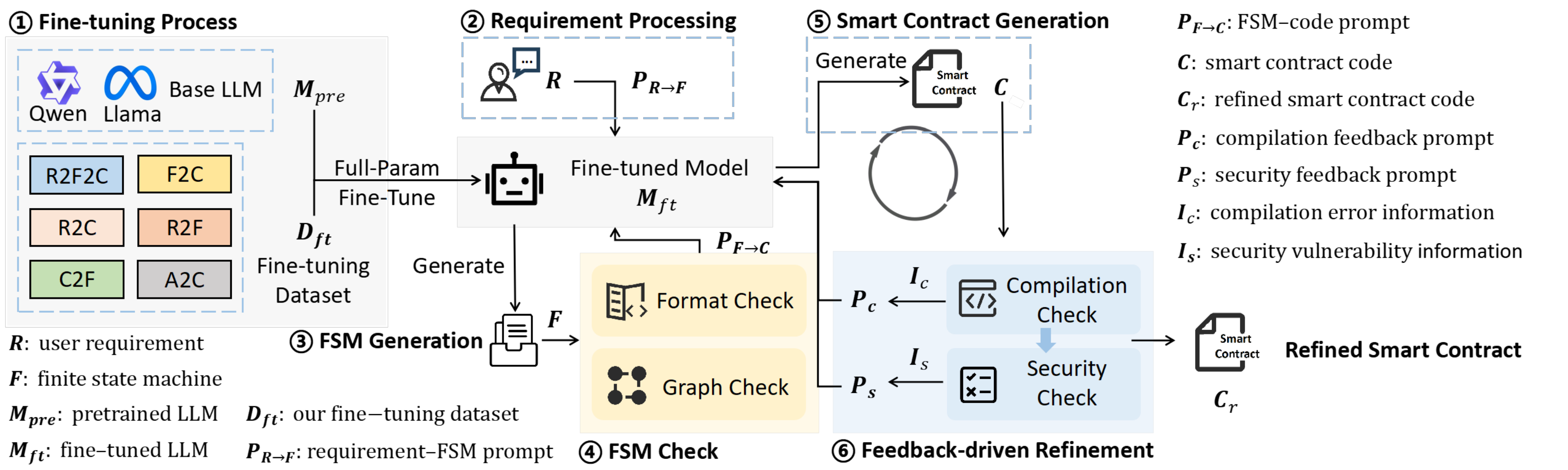}
    \centering
    
    \caption{An overview of our FSM-SCG framework.}
    \label{fig:framework}
    \vspace{-10pt}
\end{figure*}

\section{FSM Guided Smart Contract Generation}

In this part, we first provide an overview of our solution FSM-SCG. Then, we introduce its key designs, including an enhanced FSM to encode more information for smart contracts, the prompts tailored for generating FSM and smart contract, and the refinement of smart contract using feedback from compilation and security checks.   

\subsection{FSM-SCG Overview}
Figure~\ref{fig:framework} provides an overview of FSM-SCG, including the following key processes. We also provide the detailed algorithm of FSM-SCG in Appendix \ref{FSM-SCG algorithom}.

\stitle{Fine-tuning Process}
To enhance the ability of LLMs in generating smart contracts, we build a fine-tuning dataset \(D_{ft}\) that uses FSM as an intermediate representation to generate smart contracts based on user requirements (see Section~\ref{Fine-tune Dataset Construction}).
Based on this dataset, we optimize the pretrained LLM \(M_{pre}\) using the supervised full parameter fine-tuning (FPFT) method~\cite{kenton2019bert}.
In FPFT, all model parameters are updated during training, enabling the model to learn the task-specific domain. 
In this manner, the LLM can better understand user requirements, translate them into FSMs, and master the domain-specific syntax and semantics of smart contracts. Together, these adaptations significantly improve the quality of the generated code.

\stitle{Requirement processing}
This module takes the natural language descriptions of the desired smart contract from the user, where the user requirements are denoted as \(R\).
We transform \(R\) into the Requirement-FSM Prompt \(P_{R \rightarrow F}\), which can guide the LLM to generate the FSM \(F\) describing the desired smart contract.

\stitle{FSM generation}
The Requirement-FSM Prompt \(P_{R \rightarrow F}\) is fed into the fine-tuned LLM \(M_{ft}\). The \(M_{ft}\) analyzes the functionality requirements and logic within the user requirements, extracts the states and their transition conditions, and generates the corresponding FSM \(F\).

\stitle{FSM check}  
This module conducts format and graph checks on FSM \( F \). The format check ensures syntactic and content integrity, while the graph check verifies logical correctness of states and transitions. Finally, the FSM-Code Prompt \( P_{F \rightarrow C} \) is generated to produce the smart contract \( C \).

\stitle{Smart contract generation}
The checked FSM \(F\) and FSM-Code Prompt \(P_{F \rightarrow C}\) are fed into fine-tuned LLM \(M_{ft}\) to guide the generation of smart contract \(C\). Since \(F\) is rigorously checked in terms of format and logic, \(M_{ft}\) can efficiently encapsulate its states and events in smart contracts.

\stitle{Feedback Driven Refinement}  
Generated smart contracts \( C \) are evaluated for effectiveness and security. We use Py-solc-x to check for compilation errors \( I_{c} \) and Slither~\cite{feist2019slither} to detect vulnerabilities \( I_{s} \). These issues are fed back into the LLM via compilation \( P_{c} \) and security \( P_{s} \) prompts to guide refinement, producing refined smart contracts \( C_{r} \).

\subsection{Fine-tune Dataset Construction} 
\label{Fine-tune Dataset Construction}

Existing work has not provided open-source datasets~\cite{liu2024moe,zhao2024recommender,luo2024taiyi}, making it difficult to reproduce their results.
Moreover, these works consider a single task or scenario and thus cannot encompass the diverse requirements of practical smart contract generation. To tackle these two problems, we construct an open-source fine-tuning dataset that covers various tasks and adopts a dialogic format.

We collect smart contract source code from platforms like Etherscan and use GPT-4o to generate the corresponding user requirements and FSM, forming a dataset of ~30k items, each containing requirements (R), FSM (F), and code (C). From this, we derive sub-datasets by rearranging elements—\textit{R2F2C}, \textit{R2F}, \textit{F2C}, \textit{C2F}, and \textit{R2C}. Additionally, the \textit{A2C} dataset maps extracted annotations to function code. These datasets support training LLMs for various smart contract generation tasks. We present the dataset details in Appendix \ref{Fine-tuning dataset}.

\subsection{Enhanced FSM for Smart Contract}
\label{enhanced fsm}

\stitle{Limitations of Mealy}
As in Section~\ref{FSM-intro}, the Mealy machine abstracts state transitions and outputs but lacks explicit representation of functions, variables, and state-output interactions, limiting its ability to capture smart contract complexity.

\stitle{SmartFSM}
To enhance the capability of FSM to represent smart contracts, we propose \textit{SmartFSM}, an enhancement to the Mealy state machine that more accurately models the structure and functionality of smart contracts. SmartFSM divides the representation of contracts into five sections, i.e., \textit{basic information}, \textit{states}, \textit{variables}, \textit{functions}, and \textit{events}, which we introduce as follows. 

\squishlist
\item{Basic information.} This section contains the background and functional descriptions of the contract, providing an overall overview of the contract for LLM.

\item{States.} This section lists all possible states of the contract to help the LLM understand the states and transfer logic. 

\item{Variables.} This section records important data in the smart contract, serving as the conditions for state transfer. 

\item{Functions.}
Each functional module of a smart contract is usually a function that receives external inputs and may trigger state transfers. The function part resembles the input-output mechanism in Mealy state machine.

\item{Events.} This section records the key operations and state changes that occur in the contract. The triggering of each event reflects the operational state of the contract and is similar to the output function in the Mealy state machine. 
\squishend

Dividing smart contract into five sections, SmartFSM retains Mealy machine state transfer capabilities while better representing variables, functions, and underlying details. We provide an example in Appendix \ref{Appendix Generation Example}.

\subsection{Checking of SmartFSM}
\label{sec:SmartFSM check}
The SmartFSM generated by LLMs may have errors like incomplete structure or incorrect logic. Therefore, we verify its correctness through three steps: SmartFSM information extraction, format check, and graph check.

\stitle{SmartFSM information extraction}
From SmartFSM, we extract the following sets:  
The state set \( S = \{s_1, s_2, \dots, 
\\
s_n\} \), representing all possible states.  
The trigger set \( X = \{x_1, x_2, \dots, x_m\} \), representing all triggers causing state transitions.  
The target state set \( T = \{t_1, t_2, \dots, t_k\} \), where \( T \subseteq S \), representing all possible target states.  
The transition set \( \Delta = \{(s, x, t) \mid s \in S, x \in X, t \in T\} \), representing all transitions as triples.

\stitle{Format check}
This step uses static analysis methods to verify the completeness and correctness of the SmartFSM.
The correct FSM should satisfy the following conditions: 
(i) the initial state \( s_0 \) must be defined within the state set (\(s_0 \in S\)); 
(ii) the target states in transitions must be defined within the state set (\(\forall (s, x, t) \in \Delta, \ t \in S\));
(iii) the triggers must be defined within the event set (\(\forall (s, x, t) \in \Delta, \ x \in X\)).

\stitle{Graph check}
Format check ensures that the SmartFSM follows the formatting rules but it does not ensure the SmartFSM is logically correct. Graph check is used to check logical correctness as the states and state transfers of FSM can be expressed as graphs.
We first use the extracted states to construct a directed graph, where the nodes represent the states, the edges represent state transitions, and the states are connected through state transitions.
A SmartFSM can be considered correct if it satisfies the following conditions: 
(i) All states in the SmartFSM can be reached from the initial state via some path \textit{p} (\(\forall s \in S, \exists p: s_0 \to s\)); 
(ii) The state graph should have loops but no self-loops, to ensure that states can transition correctly (\(t \neq s\)).

\subsection{Prompts for FSM and Contract Generation}
\label{prompt_design}

\begin{figure}[t]
    \includegraphics[width=0.43\textwidth]{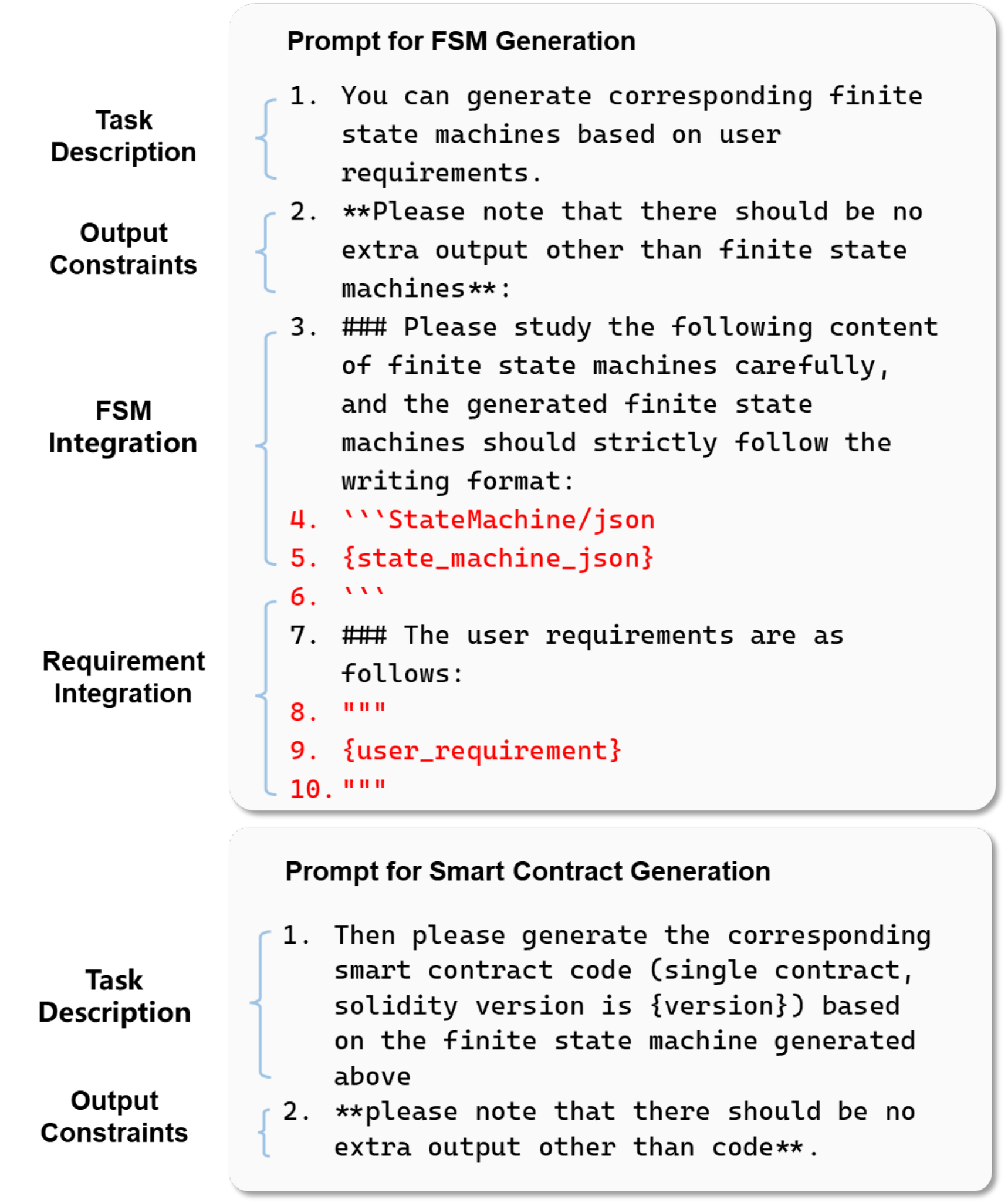}
    \centering
    \vspace{-5pt}
    \caption{The prompts for FSM and contract generation.}
    \label{fig:singlecolumn-1}
    \vspace{-10pt}
\end{figure}

Prompt engineering is crucial for effectively using LLMs.
Figure~\ref{fig:singlecolumn-1} shows the prompts for FSM and contract generation.

\stitle{Prompt for FSM generation} 
To construct FSM accurately from user requirements, we carefully design the prompt to make the LLM focus on the structure of the FSM and ensure that no unnecessary information is generated. The prompt comprises the following components:

\squishlist
    \item Task description: Guide the LLM to generate smart contracts based on user requirements.
    \item Output constraints: Request the LLM to return only the FSM content, without outputting any other information.
    \item FSM integration: The prompt explicitly provides the JSON format of the FSM. By providing a predefined format, the prompt helps the LLM understand the expected structure and syntax of FSM. This also reduces the possibility of formatting errors and deviations from the intended structure.
    
    \item Requirement integration: The prompt includes placeholders for user requirements. 
\squishend

The prompt design, by providing a task description, clearly limiting the output, and offering a structured template, enables the LLM to understand the requirements and generate reliable FSMs, while minimizing the possibility of redundant or irrelevant content.

\stitle{Prompt for smart contract generation} 
In order to generate a smart contract that meets user requirements, conforms to the specific FSM, and adheres to coding standards, we design the following prompt.

\squishlist
    \item Task description:
    As shown at the bottom of Figure \ref{fig:singlecolumn-1}, the prompt directs the LLM to generate smart contract code based on the previously generated FSM. The LLM can leverage the session context to extract the smart contract content and accurately implement the logic and structure defined in the FSM, without requiring the FSM to be included in the prompt.
    \item Output constraints: 
    Request the LLM to return the smart contract, without outputting any other information.
\squishend

The proposed prompt hence maintain logical consistency between the design phase (FSM) and the implementation phase (smart contract). 
This stratified approach ensures that the behavior represented in the FSM, such as state transitions and triggers, is reflected in the code.

\begin{figure}[!t]
    \includegraphics[width=0.43\textwidth]{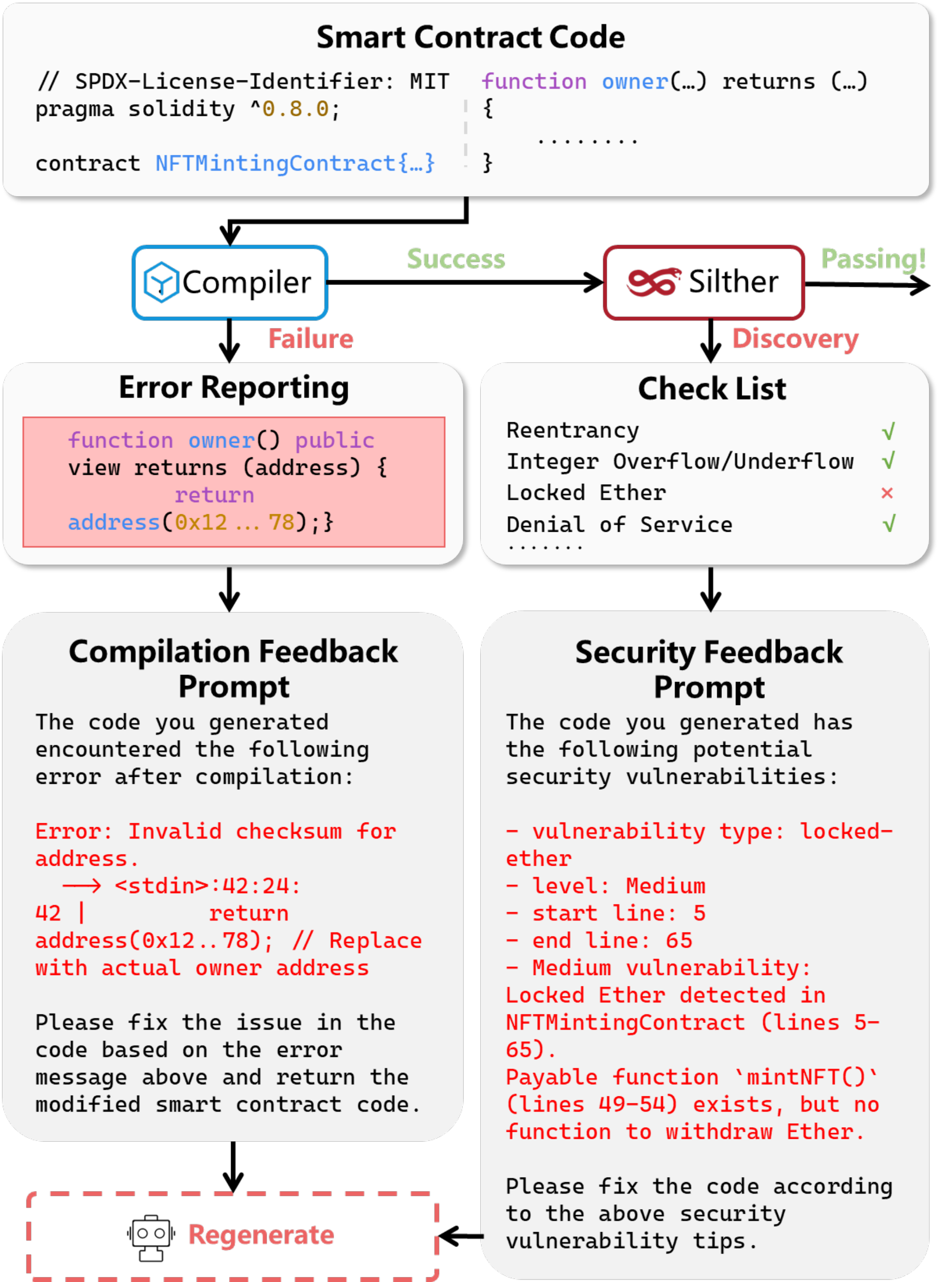}
    \centering
    \caption{Examples for compilation and security feedback.}
    \label{fig:singlecolumn-2}
    \vspace{-10pt}
\end{figure}

\subsection{Contract Refinement with Feedback}
\label{AA}

\stitle{Compilation feedback}
\label{feedback}
As shown in Figure~\ref{fig:singlecolumn-2}, we compile the generated code by Solidity compiler. If errors are detected, we return the errors to the LLM along with a contextual prompt to help improve the quality of generated code. 
In the prompt, we inform the LLM about the details of the errors and desired improvements for regeneration.
The feedback not only provides the specific reasons for failure but also offers explicit correction guidance.
As Figure~\ref{fig:singlecolumn-2} shows, during the compilation process, we identify address errors within the functions and provide the errors in the \textit{compilation feedback prompt} to guide the LLM in regenerating the contract.

\stitle{Security feedback}
To ensure code security, we use Slither to detect vulnerabilities such as reentrancy and locked assets. Detected issues, including type and code segment, are recorded in a \textit{security feedback prompt}.
Figure~\ref{fig:singlecolumn-2} shows a "Locked Ether" vulnerability, where the contract can receive but not transfer funds. Using a predefined prompt template, we integrate details like type, level, position, and reason, providing the prompt to the LLM to guide regeneration.

The above feedback procedure is iterated until compilation errors and security vulnerabilities are eliminated.

\subsection{Comparison to prior works}
\label{comparison}

Compared to the prior works \cite{napoli2024leveraging,chatterjee2024efficacy,petrovic2023model,qasse2023chat2code},
our approach has the following advantages.

\stitle{FSM as an Intermediate Representation}
Our experiments show that FSM outperforms IContractML\cite{petrovic2023model} and CML\cite{wohrer2020domain}, as it concisely captures user requirements and state transitions. In contrast, IContractML is overly complex with redundant details, while CML lacks state transition representation.

\stitle{Fine-tuning and Feedback}
We use supervised FPFT approach to enhance smart contract generation and introduce a novel feedback mechanism. Effectiveness and security checks provide feedback to improve contract quality.
\section{Experimental Evaluation}

\subsection{Experiment Settings}

\stitle{Base LLMs} 
To ensure comprehensiveness, we evaluate our method using both closed-source models (GPT-4o, GeMini1.5-Flash, Qwen-plus) and open-source models (LlaMa3.1-405B/8B, Qwen2.5-7B), covering different parameter scales.

\stitle{Baselines and Our Method}
We select three representative methods for smart contract generation as baselines.

\squishlist

\item{Direct.}
A series of research on smart contract generation uses LLMs to generate smart contract code based on user requirements directly ~\cite{napoli2024leveraging,chatterjee2024efficacy}.

\item{IContractML.}
IContractML first uses the LLMs to generate the IContractML model language, and then generates contract code from the model language~\cite{petrovic2023model,qasse2023chat2code}.
It represents methods that introduce an intermediate representation during smart contract generation.

\item {CML.} 
Contract Modeling Language (CML), a high-level DSL, has been used in traditional smart contract generation\cite{wohrer2020domain}, but not for guiding LLMs. We adopt CML as an intermediate representation and use it as a baseline for comparison.

\squishend

We compare the performance of the three variants of the methods proposed in this paper with the baseline.

\squishlist

\item{FSM.}
We employ the SmartFSM proposed in Section \ref{enhanced fsm} as an intermediate representation to guide LLMs to generate smart contracts.
However, it does not incorporate iterative feedback mechanisms or involve fine-tuning the LLM.

\item {FSM-SCG*.}
FSM-SCG* extends FSM by incorporating compilation and security feedback to iteratively optimize smart contracts, improving their effectiveness and security without fine-tuning the LLM.

\item{FSM-SCG.}
FSM-SCG further enhances FSM-SCG* by fine-tuning the LLM on our fine-tuning dataset.

\squishend

\stitle{Environment and Parameters}
We fine-tune instruction versions of LlaMa3.1-8B and Qwen2.5-7B models using 8 NVIDIA A6000-48GB GPUs. 
FPFT is conducted for 3 epochs with the AdamW optimizer, with a learning rate of 5e-5. 
In FSM-SCG, we perform one round of compilation feedback and one round of security feedback.

\begin{figure}[!t]
    \includegraphics[width=0.5\textwidth]{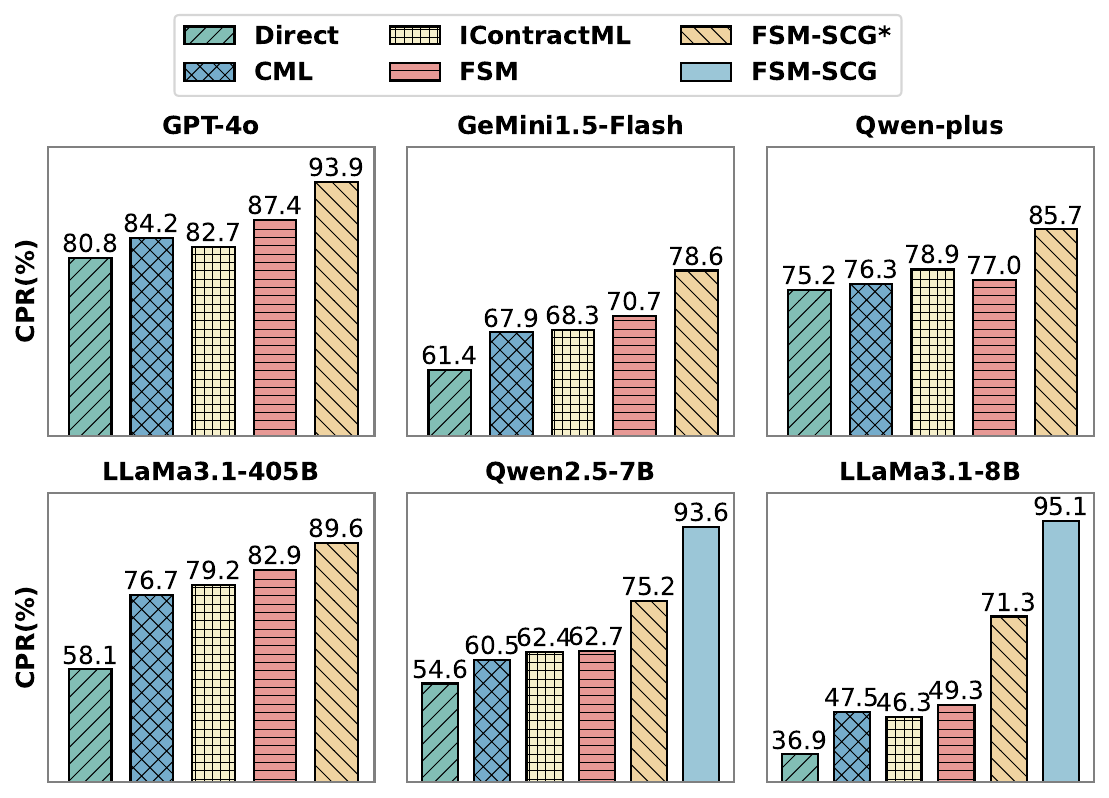}
    \centering
    \vspace{-15pt}
    \caption{Compilation Pass Rate (CPR) of smart contracts generated by each method on different LLMs. All methods generate contracts from the same 1,000 user requirements.}
    \label{fig:exp-fig-1}
    \vspace{-10pt}
\end{figure}

\subsection{Performance Metrics}
\label{sec: metric}
\stitle{CPR}
Compilation pass rate (CPR) reflects the effectiveness of the generated smart contracts.
\[
CPR = N_{\text{compiled}} / N_{\text{total}} \times 100\%
\]
\text{where} $N_{\text{compiled}}$ is the number of successfully compiled smart contracts, and $N_{\text{total}}$ is the total number of contracts.

\stitle{ZRCP and HRCP}
Zero risk contract percentage (ZRCP) and high risk contract percentage (HRCP) quantify the security of contracts.
\[
ZRCP = N_{zero} / N_{total}, \quad\quad HRCP = N_{high} / N_{total}
\]
\text{where} $N_{zero}$ is the number of contracts with zero risk scores, $N_{high}$ is the number of contracts with high-severity vulnerabilities, and $N_{total}$ is the total number of contracts.

\stitle{VRS}
Vulnerability risk score (VRS) indicates the risk level, and a high VRS may be caused by many vulnerabilities or high severity of those vulnerabilities.
\[
VRS = \frac{\sum_{i=1}^{n} (\text{SeverityScore}_i \times \text{ConfidenceScore}_i)}{n}
\]
\text{where} $n$ is the total number of detected vulnerabilities, $\text{SeverityScore}_i$ is the severity score of vulnerability $i$, which can be 3 (high), 2 (medium), or 1 (low); and $\text{ConfidenceScore}_i$ is the confidence score of vulnerability $i$,  which can be 3 (high), 2 (medium), or 1 (low). VRS is set to 10 when the code does not compile successfully.

\subsection{Main Experiments}

\stitle{Effectiveness}
We evaluate methods using the CPR of generated contracts. We sample 1,000 high-quality requirements from dataset for testing. For each requirement, LLM generates three code samples, which are compiled to compute CPR. Figure \ref{fig:exp-fig-1} presents the results.

\squishlist
    \item{\em Closed-source LLMs.}
    The FSM method uses SmartFSM as an intermediate representation to guide LLMs, outperforming Direct, CML, and IContractML in most cases on both closed-source and open-source LLMs. The FSM-SCG* method, adding generation feedback, achieves the best results among all baselines.

    \item{\em Open-source LLMs.}
    Due to fine-tuning limitations, FSM-SCG is applied only to LlaMa3.1-8B and Qwen2.5-7B, achieving significant CPR improvements, surpassing even GPT-4o. For example, the CPR of LlaMa3.1-8B rises from 36.9\% to 95.3\%, proving its effectiveness in enhancing smart contract compilation success.
    
\squishend

Our method outperforms baselines by structuring requirements into SmartFSM, covering all states and events to enhance contract generation.

\begin{table}[!tb]
\caption{Evaluation results for Security. 
We highlight the best method in \textbf{bold} and the second-best with \underline{underline}.}
\footnotesize
\setlength{\tabcolsep}{2pt}
\label{tab:exp-table-2}
\vspace{-5pt}
\centering
\begin{tabular}{ccccc}
\toprule
Model & Approach & ZRCP(↑) & HRCP(↓) & VRS(↓) \\ \hline
\multirow{5}{*}{GPT-4o} & Direct & 30.32\% & 15.47\% & 3.6720 \\ 
 & CML & 35.04\% & 12.83\% & 3.4135 \\ 
 & IContractML & 34.82\% & 13.54\% & 3.4775 \\ 
 & FSM & \underline{35.81\%} & \underline{11.90\%} & \underline{3.3235} \\
 & FSM-SCG* & \textbf{42.75\%} & \textbf{8.63\%} & \textbf{2.7649} \\ \hline
\multirow{5}{*}{GeMini1.5-Flash} & Direct & 28.90\% & 36.48\% & 4.8316 \\ 
 & CML & 28.42\% & 28.72\% & 4.8043 \\ 
 & IContractML & 29.58\% & 26.94\% & 4.7821 \\ 
 & FSM & \underline{32.69\%} & \underline{23.76\%} & \underline{4.5622} \\ 
 & FSM-SCG* & \textbf{36.90\%} & \textbf{18.58\%} & \textbf{3.9462} \\ \hline
\multirow{5}{*}{Qwen-plus} & Direct & 27.39\% & 22.74\% & 4.2730 \\ 
 & CML & 35.39\% & 23.07\% & 4.2956 \\ 
 & IContractML & 37.90\% & 20.91\% & 4.2561 \\ 
 & FSM & \underline{38.18\%} & \underline{18.12\%} & \underline{4.1063} \\
 & FSM-SCG* & \textbf{41.77\%} & \textbf{14.35\%} & \textbf{3.6983} \\ \hline
\multirow{5}{*}{LlaMa3.1-405B} & Direct & 28.22\% & 15.51\% & 6.0433 \\ 
 & CML & 30.51\% & 17.47\% & 4.4756 \\ 
 & IContractML & 29.30\% & \underline{15.28\%} & 4.1419 \\ 
 & FSM & \underline{34.39\%} & 16.65\% & \underline{3.9342} \\
 & FSM-SCG* & \textbf{39.84\%} & \textbf{13.62\%} & \textbf{3.4529} \\ \hline
 \multirow{6}{*}{Qwen2.5-7B} & Direct & 16.85\% & 22.43\% & 7.9704 \\ 
 & CML & 19.01\% & 20.33\% & 5.7987 \\ 
 & IContractML & 20.67\% & 16.03\% & 4.9844 \\ 
 & FSM & 22.01\% & 19.62\% & 5.2419 \\ 
 & FSM-SCG* & \underline{30.72\%} & \underline{14.76\%} & \underline{4.0981} \\
 & FSM-SCG & \textbf{50.53\%} & \textbf{6.62\%} & \textbf{2.4832} \\ \hline
\multirow{6}{*}{LlaMa3.1-8B} & Direct & 27.10\% & 21.68\% & 7.4416 \\ 
 & CML & 30.74\% & 32.21\% & 6.3598 \\ 
 & IContractML & 31.34\% & 37.80\% & 6.2111 \\ 
 & FSM & 33.27\% & 20.28\% & 6.0461 \\ 
 & FSM-SCG* & \underline{35.20\%} & \underline{14.87\%} & \underline{4.2540} \\
 & FSM-SCG & \textbf{53.21\%} & \textbf{5.15\%} & \textbf{2.3621} \\
\bottomrule
\end{tabular}
\vspace{-23pt}
\end{table}

\stitle{Security}
We evaluate security of each method in contract generation using ZRCP, HRCP, and VRS metrics. We sample 1,000 high-quality requirements from dataset for testing. For each requirement, LLM generates three code samples. Security is assessed through performance testing, as shown in Table \ref{tab:exp-table-2}, with the following observations:

\squishlist

\item{\em Closed-source LLMs.}
Although FSM lacks a feedback mechanism, it enhances contract security, outperforming Direct, CML, and IContractML in ZRCP and ranking second in HRCP and VRS. This shows SmartFSM improves secure contract generation, with FSM-SCG* achieving the best overall results.

\item{\em Open-source LLMs.}
After applying FSM-SCG, the percentage of smart contracts without security vulnerabilities (ZRCP) increases to 53.21\% on Qwen2.5-7B and 50.53\% on LlaMa3.1-8B, while high-risk vulnerabilities (HRCP) decrease to 5.15\% and 6.62\%, respectively.
\squishend

Our method surpasses baselines in security by ensuring valid state transitions, preventing unsafe changes, handling exceptions, and reducing vulnerabilities.

\begin{table}
\centering
\caption{Ablation studies for fine-tuning and refinement with feedback.
We highlight the best method in \textbf{bold}.}
\footnotesize
\setlength{\tabcolsep}{1pt}
\renewcommand{\arraystretch}{1} 
\label{tab:exp-table-4}
\vspace{-5pt}
\begin{tabular}{clcccc}
\toprule
Model & Approach & CPR(↑) & ZRCP(↑) & HRCP(↓) & VRS(↓) \\ \hline
\multirow{7}{*}{Qwen2.5-7B} & Direct & 54.6\% & 16.85\% & 22.43\% & 7.9704 \\ 
 & Mealy FSM & 59.8\% & 19.23\% & 25.46\% & 6.2762 \\ 
 & Our FSM & 62.7\% & 22.01\% & 19.62\% & 5.2419 \\ \cline{2-6} 
 & FSM-SCG & \textbf{93.6\%} & \textbf{50.53\%} & \textbf{6.62\%} & \textbf{2.4832} \\ 
 & - w/o Feedback & 88.5\% & 35.71\% & 14.46\% & 3.4567 \\
 & - w/o A2C Dataset & 90.9\% & 44.00\% & 8.69\% & 3.0942 \\ 
 & - w/o Fine-tuning & 75.2\% & 30.72\% & 14.76\% & 4.0981 \\ \hline
\multirow{7}{*}{LlaMa3.1-8B} & Direct & 36.9\% & 27.10\% & 21.68\% & 7.4416 \\ 
 & Mealy FSM & 37.3\% & 29.37\% & 23.03\% & 6.6347 \\ 
 & Our FSM & 49.3\% & 33.27\% & 20.28\% & 6.0461 \\ \cline{2-6} 
 & FSM-SCG & \textbf{95.1\%} & \textbf{53.21\%} & \textbf{5.15\%} & \textbf{2.3621} \\ 
 & - w/o Feedback & 91.3\% & 38.12\% & 11.94\% & 3.1061 \\
 & - w/o A2C Dataset & 92.8\% & 44.40\% & 8.73\% & 2.8349 \\ 
 & - w/o Fine-tuning & 71.3\% & 35.20\% & 14.87\% & 4.2540 \\
\bottomrule
\end{tabular}
\vspace{-5pt}
\end{table}

\subsection{Ablation Study}

We conduct ablation studies (Table \ref{tab:exp-table-4}) to assess the contributions of key components, including the feedback mechanism, A2C fine-tuning, and the overall fine-tuning process.

\squishlist

\item \textbf{Effect of SmartFSM.}
Replacing SmartFSM with Mealy FSM leads to performance degradation. As shown in the table, SmartFSM outperforms Mealy FSM w.r.t. all metrics because it better expresses user requirements and guides LLMs in generating smart contracts.

\item \textbf{Effect of feedback.}
Removing the feedback significantly reduces CPR and worsens security metrics. For example, the CPR of LlaMa3.1-8B drops from 95.1\% to 91.3\%, and VRS increases from 2.36 to 3.11. This shows that feedback improves the compilation success rate and contract security by iteratively refining the generated contracts.

\item \textbf{Effect of A2C fine-tuning.}
Removing the A2C dataset from our fine-tuning dataset and fine-tuning the LLM causes a performance drop as the A2C dataset improves LLM's ability to associate code with annotation.

\item \textbf{Effect of entire fine-tuning.}  
Without fine-tuning, all metrics degrade significantly, highlighting its importance to improve CPR and security. Fine-tuning helps the model grasp Solidity complexities, meet functional requirements, and address security vulnerabilities.

\begin{table}[!tb]
\centering
\setlength{\tabcolsep}{1pt} 
\caption{The impact of fine-tuning methods on the effect.}
\label{tab:exp-table-5}
\renewcommand{\arraystretch}{1} 
\vspace{-5pt}
\begin{tabular}{cccc}
\hline
Method & Model & CPR(↑) & VRS(↓) \\ \hline
\multirow{2}{*}{FPFT} & Qwen2.5-7B & 93.6\% & 2.48 \\ 
 & LlaMa3.1-8B & 95.1\% & 2.36 \\ \hline
\multirow{2}{*}{LoRA} & Qwen2.5-7B & 80.5\% & 3.37 \\ 
 & LlaMa3.1-8B & 75.8\% & 3.56 \\ \hline
\end{tabular}
\vspace{-10pt}
\end{table}

\item \textbf{Effect of fine-tuning method.}
We compare FPFT and LoRA \cite{hu2022lora,xia2024efficient} and include the experimental results in Table \ref{tab:exp-table-5}. FPFT outperforms LoRA by updating all parameters for better task adaptation, while LoRA’s limited updates hinder adaptability. Thus, we choose FPFT for our experiments. 
The detailed discussion is in the Appendix \ref{Parameter Sensitivity}.

\squishend

\subsection{Parameter Sensitivity}
We also conduct experiments to evaluate the parameter sensitivity of FSM-SCG (see Appendix \ref{Parameter Sensitivity}). 
Specifically, we investigate the impact of feedback counts and fine-tuning epochs.

\squishlist

\item \textbf{Effect of feedback count.}  
Varying the feedback count from 0 to 5, the largest improvement occurs from 0 to 1, as most errors are simple and need minimal iterations. Thus, we set the feedback count to 1 for a balance of performance and efficiency.

\item \textbf{Effect of fine-tuning epochs.}  
We fine-tune FSM-SCG from 1 to 5 epochs. Performance improves notably up to 3 epochs but plateaus thereafter, as the model converges early and saturates with further training. To balance efficiency and performance, we select 3 epochs.

\squishend

\section{Related Work}

\stitle{Automatic Generation Methods for Smart Contracts}
Smart contract generation methods include Model-Driven generation (MD)\cite{lopez2017caterpillar,tran2018lorikeet}
, Domain-Specific Languages (DSL)\cite{clack2019smart,skotnica2019contract}
, and Visual Programming Languages (VPL)\cite{mavridou2018tool,mao2019visual}.
MD enhances design-code consistency by converting domain models into smart contracts but requires specialized knowledge. 
DSL simplifies contract development within specific domains but lacks broader applicability and robust verification. 
VPL uses graphical interfaces to lower development barriers but offers limited functionality and scalability.

\stitle{Generation of Smart Contracts Using LLMs}
The development of LLMs has made automatic smart contract generation a research focus\cite{yang2024harnessing}. 
Chatterjee et al.\cite{chatterjee2024efficacy} explored using descriptive and structured prompts to generate smart contracts, finding quality issues and neglect of security by most LLMs. 
Petrović and Al-Azzoni\cite{petrovic2023model} proposed a model-driven framework using ChatGPT, but it suffers from long response times and high costs.
Chen et al.\cite{yong2024smart} improved contract quality by combining AST-LSTM representation, clustering models, prompt dataset optimization, and fine-tuning LLaMA2-7B. 
Qasse et al.\cite{qasse2023chat2code} extended DSLs with chatbots, improving functionality but facing challenges with input errors and accuracy.
However, LLMs still face issues like incorrect syntax, security vulnerabilities, and limited adaptability in this domain \cite{huang2024retrofitting,huang2024learning}.

\stitle{Anatomy of Prior Works}
Most prior works, except \cite{chatterjee2024efficacy}, rely on a single LLM for smart contract generation, and none has fine-tuned LLMs for this task. Analyzing a single LLM is insufficient, therefore, our work employs six LLMs within the FSM-SCG framework. We assess contract generation through compilation checks, security analysis, and use case testing. A feedback mechanism further enhances trustworthiness by guiding LLMs to refine output, distinguishing our approach and improving performance \cite{zhang2024treecss,wang2024self}.
\section{Conclusion}


In this work, we propose an FSM-guided framework for smart contract generation with LLMs, improving effectiveness and security. We release a fine-tuning dataset to better align LLMs with user needs and integrate compilation and security checks into a feedback mechanism for improved reliability. 
\newpage
\section*{Acknowledgments}

This work was sponsored by National Key R\&D Program of China (No.2023YFB2703600), Key R\&D Program of Hubei Province (2023BAB077), the Fundamental Research Funds for the Central Universities (2042025kf0040), and National Natural Science Foundation of China (62472327, 62441229). This work was supported by Sichuan Clinical Research Center for Imaging Medicine (YXYX2402).

\section*{Contribution Statement}
Hao Luo and Yuhao Lin contributed equally to this work.

\bibliographystyle{named}
\bibliography{ijcai25}

\clearpage
\appendix

\begin{center}
\bf \huge Technical Appendix
\vspace{2mm}
\end{center}

\section{FSM-SCG}
\label{FSM-SCG algorithom}

\begin{algorithm}[!h]
\caption{FSM-SCG Algorithm Workflow}
\label{algorithom workflow}
\quad \textbf{Input:} User requirements \( R \)\par
\quad \textbf{Output:} Refined smart contract \( C_r \)\par
\quad \text{Refer to Fig.~\ref{fig:framework} for other notations.} 
\begin{algorithmic}[1]
\STATE \( M_{ft} \leftarrow \text{FPFT}(M_{pre}, D_{ft}) \) 
\STATE \(P_{R \to F} \leftarrow \text{R2FPrompt}(R)\)
\STATE \( F \leftarrow M_{ft}(P_{R \to F}) \) 
\WHILE{\NOT (\text{FormatCK}(\(F\)) \AND \text{GraphCK}(\(F\)))}
    \STATE \( F \leftarrow M_{ft}(R) \)
\ENDWHILE
\STATE \(P_{F \to C} \leftarrow \text{F2CPrompt}(F)\)
\STATE \( C \leftarrow M_{ft}(P_{F \to C}) \)
\STATE \(SuccessFlag \leftarrow \text{False}\)
\WHILE{\NOT \(SuccessFlag\)}
    \IF{\NOT \text{CompileCK}(\(C\))}
        \STATE \(P_{c} \leftarrow \text{CompilePrompt}(I_c)\)
        \STATE \( C \leftarrow M_{ft}(P_{c}) \)
    \ELSE
        \IF{\NOT \text{SecurityCK}(\(C\))}
            \STATE \(P_{s} \leftarrow \text{SecurityPrompt}(I_s)\)
            \STATE \( C \leftarrow M_{ft}(P_{s}) \)
        \ELSE
            \STATE \(SuccessFlag \leftarrow \text{True}\)
        \ENDIF
    \ENDIF
\ENDWHILE
\STATE \( C_r \leftarrow C \)
\RETURN \( C_r \)
\end{algorithmic}
\end{algorithm}
\vspace{-5pt}

The FSM-SCG algorithm fine-tunes a pre-trained model \( M_{ft} \) using a dataset \( D_{ft} \), generates a Finite State Machine (FSM) \( F \) from user requirements \( R \) via a prompt \( P_{R \to F} \), and validates \( F \) through format (\texttt{FormatCK}) and graph (\texttt{GraphCK}) checks in a loop until it passes. It then generates a smart contract \( C \) from \( F \) using a prompt \( P_{F \to C} \), and iteratively refines \( C \) by re-generating it with compilation (\texttt{CompilePrompt}) or security prompts (\texttt{SecurityPrompt}) until it passes both compilation (\texttt{CompileCK}) and security (\texttt{SecurityCK}) checks. Finally, the refined smart contract \( C_r \) is returned after performing unit tests using a dataset \( D_b \).

Once a valid FSM \( F \) is obtained, the algorithm generates a smart contract \( C \) by creating a prompt \( P_{F \to C} \) using the \texttt{F2CPrompt} function and processing it through \( M_{ft} \). The smart contract \( C \) undergoes further refinement through iterative validation loops. If \( C \) fails the compilation check (\texttt{CompileCK}), a compilation prompt \( P_{c} \) is generated using \texttt{CompilePrompt}, and \( M_{ft} \) regenerates \( C \). Similarly, if \( C \) fails the security check (\texttt{SecurityCK}), a security prompt \( P_{s} \) is generated using \texttt{SecurityPrompt}, and \( M_{ft} \) regenerates \( C \). This process continues until \( C \) passes both compilation and security checks.

Finally, the refined smart contract \( C_r \) is set to the validated \( C \) and returned as the final output, ensuring that it can pass compilation and meets security standards.

\section{Fine-tuning Dataset}
\label{Fine-tuning dataset}
We construct a dataset to fine-tune LLMs for smart contract generation.

\stitle{Motivation and overview} Fine-tuning has been used by existing methods to enhance the performance of LLMs for smart contract generation~\cite{liu2024moe,zhao2024recommender,luo2024taiyi}. 
However, these datasets are not public, which makes it difficult to reproduce their results.
Moreover, these datasets consider a single task or scenario and thus cannot encompass the diverse requirements of practical smart contract generation. To tackle these two problems, we construct an open-source fine-tuning dataset that covers various tasks and adopts a dialogic format.

We collect smart contract source code from platforms such as Etherscan and use GPT-4o to generate the corresponding user requirements and FSM, resulting in a main dataset of approximately 30k data items.
Each item consists of user requirements (R), FSM (F), and smart contract code (C).
Based on this main dataset, we produce the following sub-datasets by rearranging the elements of the items. These sub-datasets can be used to train models for different tasks during smart contract generation. 

\squishlist
\item{\textit{R2F2C dataset}}
 contains the R, F, and C. It can be used to train models that generate FSMs from user requirements and transform FSMs into smart contracts.

\item{\textit{R2F dataset}}
contains the R and F. We train models to map user requirements to FSMs.

\item{\textit{F2C dataset}}
contains F and C, and can be used to train models to transform FSMs into smart contracts.

\item{\textit{C2F dataset}}
contains F and C. It allows models to derive FSMs from smart contract codes and thus understand the connection between FSM logic and contract code.

\item{\textit{R2C dataset}}
contains R and C. We train models to directly map user requirements to smart contract codes.

\item{\textit{A2C dataset}}
contains annotation extracted from smart contracts, along with the corresponding function code. It can be used to train models that generate function code based on these annotations.
\squishend

\begin{figure*}[!t]
    \centering
    \includegraphics[width=0.8\textwidth]{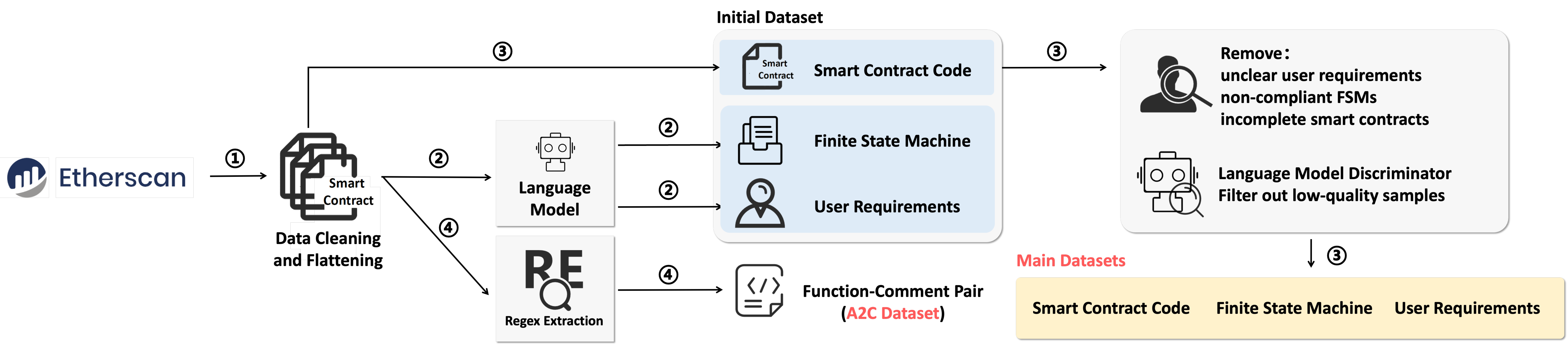}
    \caption{Procedure for generating the fine-tuning dataset.}
    \label{fig:dataset generation}
    \vspace{-5pt}
\end{figure*}

\label{Generation example}
\begin{figure*}[!t]
    \includegraphics[width=0.8\textwidth]{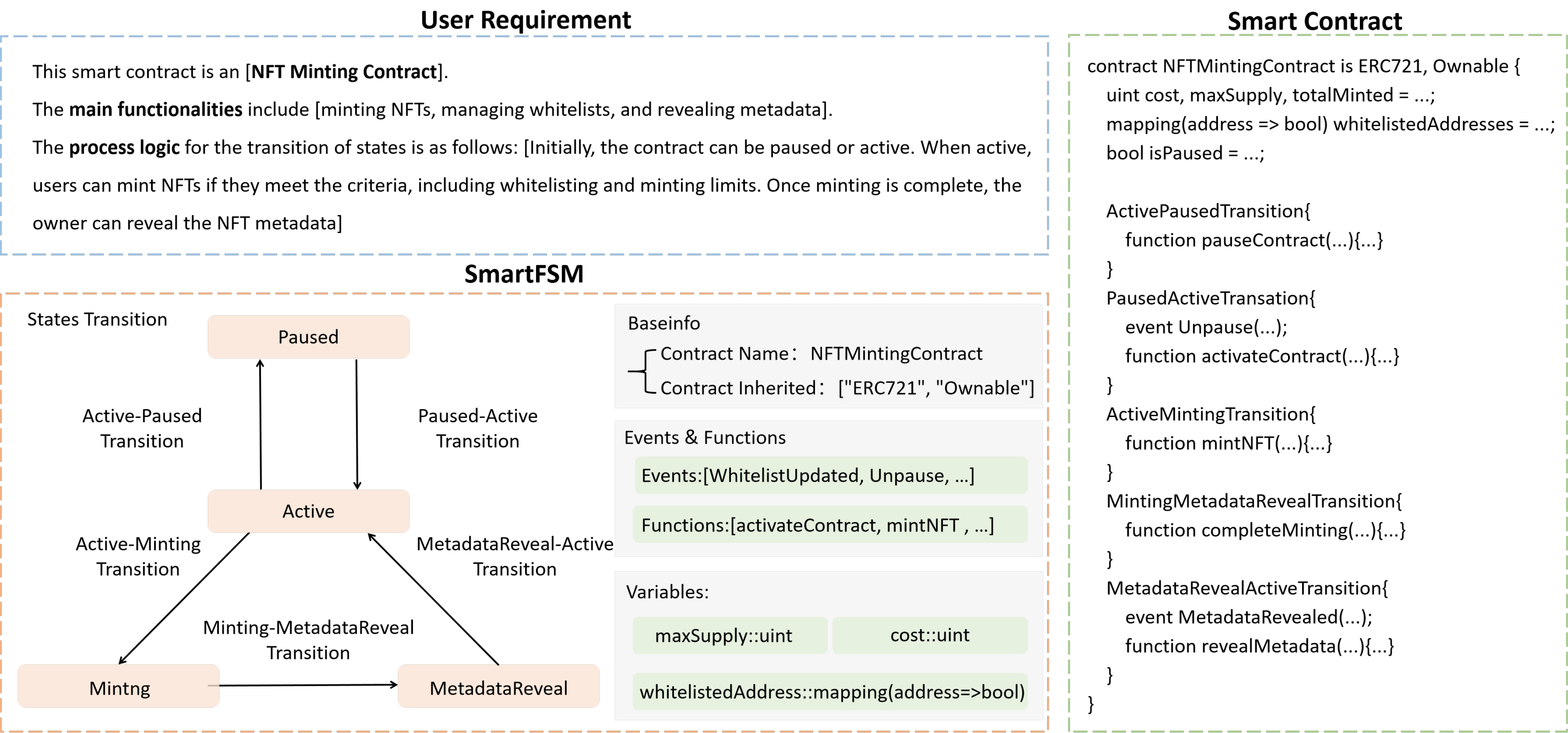}
    \centering
    \caption{Three key elements involved in the smart contract generation process: Requirement, FSM, and Smart Contract.
    }
    \label{fig:three key elements}
    \vspace{-10pt}
\end{figure*}

\stitle{Generation procedure} As shown in Figure~\ref{fig:dataset generation}, it takes 4 steps to generate the fine-tuning dataset.

\stitle{\ding{182} Data collecting and cleaning}
We download smart contracts from sources like Etherscan and collect contract codes with varying functionalities, language versions, and complexities, which serve as raw material for constructing the dataset. For highly similar smart contracts, we retain the most complete and representative samples.

\stitle{\ding{183} Extracting FSM and requirements}
We input the preprocessed smart contract code into the GPT-4o and generate the corresponding FSM and user requirements.

\stitle{\ding{184} Constructing the main dataset}
We collect the code (C), FSM (F), and user requirements (R) of each smart contract to form the initial dataset.
After manual inspection and GPT-4o scoring, the initial dataset is refined into the main dataset, with inaccurate descriptions, irregular FSMs, and incomplete smart contracts removed.
Based on this main dataset, the R2F, F2C, R2F2C, C2F, and R2C subdatasets can be generated.

\stitle{\ding{185} Annotation extraction}
We scan the preprocessed smart contract code with regular expressions for pairs of functions and annotation. This produces the A2C dataset.


\section{Generation Example}
\label{Appendix Generation Example}
As shown in Figure \ref{fig:three key elements}, we use a specific example from the non-fungible token (NFT) minting scenario to illustrate the key elements involved in the smart contract generation process: user requirements, FSMs, and smart contract code.

\section{Additional Experiments}
\label{Parameter Sensitivity}

\begin{table}[htb]
\centering
\setlength{\tabcolsep}{1pt} 
\vspace{-8pt}
\caption{The impact of fine-tuning methods on the effect.}
\label{tab:exp-table-6}
\renewcommand{\arraystretch}{1} 
\begin{tabular}{cccc}
\hline
Method & Model & CPR(↑) & VRS(↓) \\ \hline
\multirow{2}{*}{FPFT} & Qwen2.5-7B & 93.6\% & 2.48 \\ 
 & LlaMa3.1-8B & 95.1\% & 2.36 \\ \hline
\multirow{2}{*}{LoRA} & Qwen2.5-7B & 80.5\% & 3.37 \\ 
 & LlaMa3.1-8B & 75.8\% & 3.56 \\ \hline
\end{tabular}
\vspace{-10pt}
\end{table}

\begin{figure}[!tb]
    \includegraphics[width=0.45\textwidth]{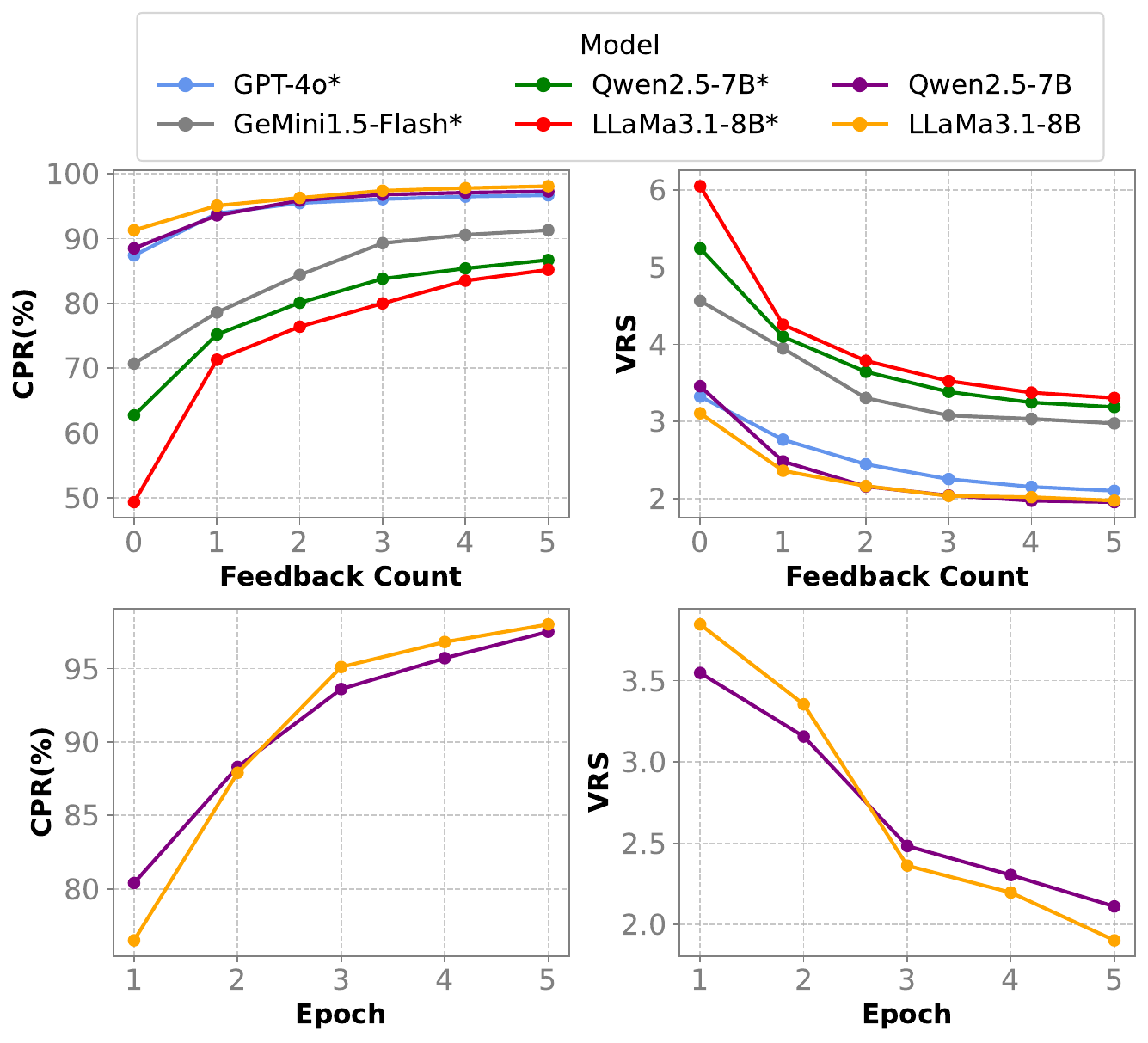}
    \centering
    \caption{Comparison of model performance with respect to feedback count and training epochs. The asterisk (*) indicates that we do not fine-tune the model.}
    \label{fig:feedback count}
    \vspace{-10pt}
\end{figure}

\stitle{Effect of fine-tuning method}
As shown in Table \ref{tab:exp-table-5}, we compare FPFT and LoRA fine-tuning while fixing all other parameters. 
FPFT outperforms LoRA as it updates all parameters, allowing better adaptation to the target task. 
In contrast, LoRA freezes model weights and updates fewer parameters, which limits its adaptability to specific tasks. 
Moreover, since the base model has not been exposed to our target task (FSM as an intermediate representation for smart contract code generation), LoRA's more limited parameter updates result in insufficient depth in task-specific learning during training. Therefore, we choose FPFT for our experiments.

\stitle{Effect of feedback count}
As shown in Figure \ref{fig:feedback count}, we vary the feedback count of FSM-SCG from 0 to 5. 
When the count increases from 0 to 1, we observe the largest performance improvement.
This is because most errors in the generated contracts are simple, requiring only a few iterations to correct them.
In the previous experiment, we choose a feedback count of one to balance performance gain and time consumption.

\stitle{Effect of fine-tuning epochs}
As shown in Figure \ref{fig:feedback count}, we vary the fine-tuning epochs of FSM-SCG from 1 to 5. 
The performance improves significantly as the epoch number increases from 1 to 3, but the gain tends to diminish beyond 3.
This is because, during the initial stages of fine-tuning, the model can quickly adapt to the new task and data, leading to substantial model convergence.
As training progresses, the model's learning gradually saturates, leaving less room for further improvements. 
To balance training time and performance, we choose 3 epochs in the experiments.

\end{CJK}
\end{document}